\documentclass[letterpaper, 10 pt, conference]{ieeeconf}  %
\IEEEoverridecommandlockouts                              %
\usepackage[hidelinks]{hyperref}
\usepackage{graphicx} 
\usepackage{multirow}
\usepackage{svg}
\usepackage{multicol}
\usepackage{float}
\usepackage{caption}
\usepackage{subcaption}
\usepackage{bbm}
\usepackage{amsmath} 
\usepackage{amssymb}  
\usepackage{xcolor}
\usepackage{soul}
\usepackage{booktabs}
\usepackage{csquotes}
\definecolor{ld_blue}{rgb}{0, 0, 0.6}

\definecolor{ld_green}{rgb}{0, 0.6, 0}

\newcommand{\norm}[1]{\left\lVert#1\right\rVert}

\title{\LARGE \bf Constrained Reinforcement Learning for Unstable Point-Feet Bipedal Locomotion Applied to the Bolt Robot}

\author{Constant Roux$^{1}$, Elliot Chane-Sane$^{1}$, Ludovic De Matteïs$^{1}$, Thomas Flayols$^{1}$, \\ Jérôme Manhes$^{1}$, Olivier Stasse$^{1,2}$ and Philippe Souères$^{1}$
\thanks{$^{1}$LAAS-CNRS, Universit\'e de Toulouse, France, {firstname.surname@laas.fr}}%
\thanks{$^{2}$Artificial and Natural Intelligence Toulouse Institute, France, {firstname.surname@laas.fr}}}%

\begin{document}

\makeatletter
\let\@oldmaketitle\@maketitle%
\renewcommand{\@maketitle}{\@oldmaketitle
	\begin{center}
		\includegraphics[width=\linewidth]{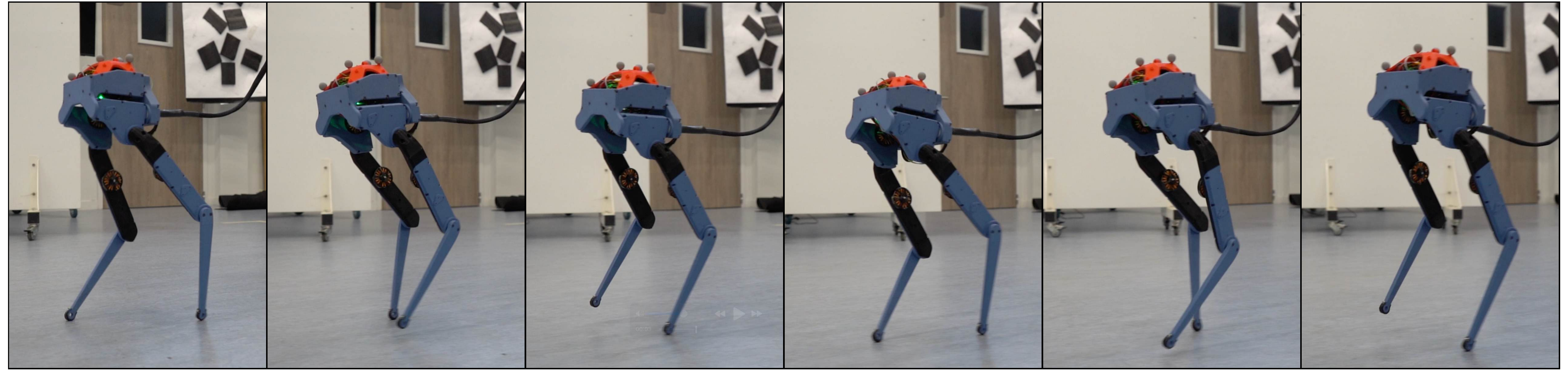}
		\captionof{figure}{The Bolt robot walking forward.}
		\label{fig:presentation}
	\end{center}}
\makeatother

\maketitle
\thispagestyle{empty}
\pagestyle{empty}



\begin{abstract}
Bipedal locomotion is a key challenge in robotics, particularly for robots like Bolt, which have a point-foot design. 
This study explores the control of such underactuated robots using constrained reinforcement learning, addressing their inherent instability, lack of arms, and limited foot actuation. 
We present a methodology that leverages Constraints-as-Terminations and domain randomization techniques to enable sim-to-real transfer.
Through a series of qualitative and quantitative experiments, we evaluate our approach in terms of balance maintenance, velocity control, and responses to slip and push disturbances.
Additionally, we analyze autonomy through metrics like the cost of transport and ground reaction force. 
Our method advances robust control strategies for point-foot bipedal robots, offering insights into broader locomotion.
Videos and code are available at \url{https://gepetto.github.io/BoltLocomotion/}.
\end{abstract}

\section{Introduction}
Traditionally, Model Predictive Control (MPC) has been the dominant method for controlling legged robots \cite{10286076, cafe2025}.
However, the emergence of advanced Reinforcement Learning (RL) algorithms \cite{schulman2017proximal} and the development of GPU-accelerated simulators \cite{mittal2023orbit,todorov2012mujoco} have driven a significant shift toward learning-based policies \cite{reviewmajid}, especially in the context of quadrupedal locomotion \cite{pmlr-v164-rudin22a, Ji2022}. 
In recent years, such techniques have also demonstrated great success in controlling bipedal and humanoid robots \cite{2406.10759, 2401.16889, 2303.03381}, enabling these robots to perform in increasingly complex environments and often outperforming traditional MPC-based approaches.

Despite these advancements, sim-to-real transfer remains challenging for robots with unconventional morphological traits, such as limited ground contact or underactuated limbs \cite{10610069, 2107.04034, 2502.01143, 10.1115/1.4055770, 10670293}. 
Among these, bipedal robots with point feet \cite{4307016, LimXDynamics_TRON1, 1234651, pfhumanoid} pose a unique challenge due to their inherently unstable structure and minimal support area. 
While some prior works have achieved point-foot bipedal walking using quadrupedal robots \cite{cheng2023parkour, 10611442}, these systems benefit from additional limbs that contribute to balance through angular momentum \cite{Gao2023}—a strategy unavailable to arm-less bipedal robots.
The scientific interest in point-foot bipedal robots is considerable. 

As noted by Westervelt et al., \textquote{\textit{a model of walking with a point contact is an integral part of an overall model of walking that is more anthropomorphic in nature than the current flat-footed walking paradigm}} \cite{Westervelt2018}. 
Unlike rigid flat-footed robots—which often adopt overly simplified and non-anthropomorphic designs with stiff, unarticulated contact surfaces—point-foot systems better reflect the dynamics and control challenges seen in human locomotion. 
To explore this issue, we focus on the point-foot Bolt robot \cite{Grimminger2020}, a bipedal platform developed as part of the Open Dynamic Robot Initiative (ODRI). 
Bolt shares the same actuators as Solo, a quadrupedal robot from the same project, leveraging a modular and open-source design aimed at advancing torque-controlled legged locomotion.
Notably, while Solo has been extensively studied \cite{chane2024cat, Aractingi2023, chane2024rlwav}, there has been relatively little research on Bolt \cite{9361296}, making it an interesting plateform for our subject of interest.

In this paper, we address the control of bipedal point-foot robots without arms, using the Bolt platform as a representative system. 
We analyze its dynamics and identify key challenges that must be tackled to enable robust locomotion.
Our contributions are as follows:
\begin{itemize}
    \item \textbf{First Application of Constrained RL on Point-Foot Bipedal Locomotion:} We apply constrained reinforcement learning to the problem of point-foot bipedal locomotion, demonstrating its feasibility on a setting that has received limited attention in prior work.
    \item \textbf{Open-Source Training and Inference Pipeline:} We provide a fully open-source pipeline for training, inference, and logging on the low-cost, open-source Bolt robot, aiming to enhance reproducibility and support future benchmarking efforts.
    \item \textbf{Real-World Evaluation:} We conduct a real-world evaluation of the approach on the Bolt robot, providing insights into its performance and practical deployment.
\end{itemize}

The structure of the paper is as follows: Section \ref{sec:related_work} reviews related work on constrained RL and point-foot robots, Section \ref{sec:hardware} provides an in-depth description of the Bolt hardware, Section \ref{sec:method} outlines the methodology, Section \ref{sec:experiments} presents experimental results, and Section \ref{sec:conclusion} concludes the paper.
\section{Related Work}
\label{sec:related_work}

Recent advances in GPU-accelerated simulators \cite{mittal2023orbit, todorov2012mujoco} have enhanced the efficiency of RL for robotic locomotion by enabling massively parallel training, which reduces training times \cite{pmlr-v164-rudin22a}. 
RL has been successful in quadrupedal locomotion, with policies trained in simulation effectively transferring to real-world robots \cite{Zhang2024, Ji2022, Aractingi2023, cheng2023parkour}. 
However, RL for bipedal locomotion remains more challenging due to the inherent instability of bipedal morphology \cite{2406.10759, 2401.16889, 2303.03381}, especially in point-foot bipedal robots, where the underactuated stance phase complicates the control \cite{10611442}. 
The absence of arms further increases the challenge by eliminating mechanisms to counteract angular momentum \cite{LimXDynamics_TRON1, li2019using, 10569443}. 
The Bolt robot, a point-foot biped, embodies many of the fundamental challenges in dynamic legged locomotion. 
Previous work has leveraged MPC and whole-body MPC to achieve stable walking behaviors \cite{9659392, 9361296, 10769884}. 
In this work, we instead focus on RL to develop more robust and adaptable locomotion policies that can better handle uncertainty and disturbances. 

Transferring RL policies to real-world bipedal robots is hindered by issues like unmodeled dynamics, sensor noise, and hardware limitations \cite{2406.10759, 2401.16889, 2303.03381}. 
Constrained RL frameworks enhance safety and robustness by incorporating constraints, such as joint position and torque limits, as termination conditions \cite{chane2024cat, chanesane2024soloparkour, chane2024rlwav}. 
One notable approach, \textit{CaT (Constraints as Terminations)}, enforces these constraints during training to ensure hardware-compliant locomotion \cite{chane2024cat}. 
Alternative methods, such as \cite{10530429}, address similar challenges through different constraint-handling techniques.
Furthermore, domain randomization techniques \cite{10610069} expose policies to various training dynamics, such as variations in motor gains and friction coefficients, enhancing robustness to real-world discrepancies. 
Additionally, adaptive control strategies allow policies to adapt to real-world conditions by addressing discrepancies between simulation and reality \cite{2107.04034, 2502.01143, fey2025bridging}. 
In this work, we employ the \textit{CaT} framework along with domain randomization to bridge the sim-to-real gap for the Bolt robot, allowing us to develop robust locomotion policies that account for real-world uncertainties.

Evaluation metrics for bipedal locomotion encompass multiple aspects of performance. 
Velocity tracking accuracy assesses how well the robot follows commanded velocity profiles, ensuring precise motion control \cite{10268037}. 
Maximum achievable speed evaluates the robot's top velocity, reflecting its control limits \cite{doi:10.1163/156855390X00305, 8594448}.
Autonomy is quantified through the cost of transport (CoT), which measures energy efficiency by calculating the energy consumed per unit distance traveled \cite{Luneckas2021, stasse:frontiers:2018}. 
Additionally, ground reaction force (GRF) profiles provide insights into balance, impact absorption, and adherence to physical constraints by analyzing force distribution during locomotion \cite{5645365, 7254255}.
Robustness is assessed through push recovery, which tests the robot's ability to regain stability after external perturbations \cite{9420230, dantec:hal-04647996, 9082021}, and slippage recovery, which evaluates its ability to maintain balance when encountering unexpected loss of traction \cite{10268037, 9082021}.
In this work, we comprehensively evaluate the Bolt robot's RL policies across all these metrics, analyzing its velocity tracking accuracy, maximum speed, energy efficiency, and robustness to external disturbances. 
This assessment offers insights into the behavior of the locomotion framework in real-world scenarios.

\section{Hardware}
\label{sec:hardware}

\subsection{Bolt Overview}
\label{sec:bolt_overview}

Bolt (as shown in Fig.~\ref{fig:presentation}) is a bipedal robot developed by the ODRI \cite{Grimminger2020}. 
It adopts a fully open-source, modular design philosophy aimed at enhancing accessibility and reproducibility in robotic research. 
Built with low-cost, off-the-shelf components and 3D-printed parts, Bolt features three torque-controlled degrees of freedom per leg and point feet. 
The robot omits arms entirely, which contributes to its lightweight form factor. 

The Bolt robot relies on an external system, connected via a wired interface, for high-level computation and power supply.
Commands are transmitted in real time via an Ethernet link to low-level motor controllers, enabling agile, closed-loop control while minimizing onboard complexity.

\subsection{Locomotion Challenges}
\label{sec:bolt_challenges}

While Bolt's streamlined design facilitates experimentation, it introduces several critical challenges for dynamic locomotion:

\begin{itemize}
    \item \textbf{Limited Stability:} Point feet provide no flat contact surface, significantly reducing the support polygon and making balance control more demanding, particularly during dynamic motions or on uneven terrain \cite{1234651}.

    \item \textbf{Angular Momentum Regulation:} The absence of arms limits the robot's ability to counterbalance body motion, placing a greater burden on the legs and torso to manage angular momentum during gait transitions and external disturbances \cite{Gao2023}.

    \item \textbf{Restricted Yaw Control:} Without an active yaw mechanism, Bolt struggles to reorient itself efficiently, which complicates tasks such as turning in confined spaces or navigating sharp trajectories \cite{Yaw2012}.

    \item \textbf{Reduced Kinematic Redundancy:} Each leg's three actuators limit the robot's motion versatility, making agile behaviors such as running or adaptive stepping more complex to achieve and requiring sophisticated control strategies.
\end{itemize}
\section{Method}
\label{sec:method}
In order to train a locomotion policy for the point-foot robot, Bolt, we use a sim-to-real approach. 
We first train the policy in simulation with deep RL and then directly transfer it to the real robot. 
This section describes the main components used to support learning and sim-to-real transfer on the Bolt robot.

\subsection{Constrained Reinforcement Learning}
Consider a Constrained Markov Decision Process defined as \( (\mathcal{S}, \mathcal{A}, \mathcal{R}, \gamma, \mathcal{T}, \{\mathcal{C}^i\}_{i \in I}) \), where \( \mathcal{S} \) and \( \mathcal{A} \) represent the state and action spaces, respectively, $\gamma$ is the discount factor, $\mathcal{R}: \mathcal{S} \times \mathcal{A} \to \mathbb{R}^+$ is the reward function, and $\mathcal{T}: \mathcal{S} \times \mathcal{A} \times \mathcal{S} \to \mathbb{R}^+$ defines the probabilistic transition dynamics.
The system is subject to constraints $\{\mathcal{C}^i: \mathcal{S} \times \mathcal{A} \to \mathbb{R}\}_{i \in I}$, where each constraint \( \mathcal{C}^i \) yields a scalar penalty signal \( c^i \in \mathbb{R}^+ \) for a given state-action pair \( (s, a) \).
We look for a policy $\pi: \mathcal{S} \to \mathcal{A}$ that maximizes the discounted sum of future rewards:
\begin{equation}
    \max _\pi \mathbb{E}_{\tau \sim \pi, \mathcal{T}}\left[\sum_{t=0}^{\infty} \gamma^t r\left(s_t, a_t\right)\right],
	\label{eq:mdp}
\end{equation}
while ensuring the constraints satisfaction:
\begin{equation}
	\mathbb{P}_{(s, a) \sim \rho^{\pi, \mathcal{T}}_\gamma} \left [ c^i(s, a) > 0 \right ] \leq \epsilon_i, \quad \forall i \in I.
	\label{eq:cstr_prob}
\end{equation}
We solve the problem defined by \eqref{eq:mdp} and \eqref{eq:cstr_prob} using the \textit{CaT} framework \cite{chane2024cat}, built on Proximal Policy Optimization (PPO) \cite{schulman2017proximal}, which incorporates constraints as termination conditions during training.

\subsection{States and Actions}
The state vector \( s \in \mathcal{S} \) comprises the commanded base velocity \(\begin{pmatrix} \mathbf{v}^{*\top} & \omega^* \end{pmatrix}^\top\), where \( \mathbf{v}^* \) and \( \omega^* \) represents respectively the desired linear and angular velocities in the robot's base frame.  
It also includes the base angular velocity \( \omega \), the projected gravity, the joint positions \( \mathbf{q} \), the joint velocities \( \dot{\mathbf{q}} \), and the previous actions.  
The six-dimensional action space corresponds to the angular position commands for the Bolt robot’s joints. 
The policy outputs scaled delta actions, which are added to a predefined reference angular configuration and then tracked by a high-frequency proportional-derivative (PD) controller.

\subsection{Rewards}
The reward function is designed to promote precise tracking of the commanded base velocity, with separate terms for tracking the linear velocity in the \(xy\)-plane and the yaw velocity.
The reward of the linear velocity is defined as:
\begin{equation}
	R_{xy}(s) = \exp\left( - \frac{\norm{\mathbf{v}^* - \mathbf{v}}^2}{\sigma_1^2} \right)
\end{equation}
where $\mathbf{v}$ denote the measured linear velocity expressed in the robot's base frame, and $\sigma_1$ represents a scaling factor.

Similarly, the reward for the yaw velocity is defined as:
\begin{equation}
	R_{\text{yaw}} = \exp\left( - \frac{( \omega^* - \omega )^2}{\sigma_2^2} \right)
\end{equation}
where $\sigma_2$ represents a scaling factor.

\subsection{Constraints}
\label{subsec:constraints}
To ensure safe and transferable locomotion, we define three categories of constraints during training: \textbf{safety constraints}, \textbf{posture constraints}, and \textbf{gait constraints}. 
These are detailed in Table~\ref{tab:constraints-bolt-rl}.

\paragraph{Safety Constraints}  
These constraints prevent critical failures that could damage the robot. 
They include prohibiting knee or base collisions with the ground, avoiding upside-down states, and limiting excessive foot contact forces \(f^{\text{foot}_j}\) of foot \(j\) with a threshold \(f^{\text{lim}}\). 
Actuator safety is ensured by limiting joint torques \( \tau_k \), joint velocities \( \dot{q}_k \), and accelerations \( \ddot{q}_k \), where \(k\) denotes the actuator index. These are constrained within limits \(\tau^{\text{lim}}\), \(\dot{q}^{\text{lim}}\), and \(\ddot{q}^{\text{lim}}\), respectively.

\paragraph{Posture Constraints}  
Posture constraints regulate the robot's body configuration. 
The base orientation in the roll and pitch axes is constrained to prevent excessive tilting, with a limit defined by \(\text{base}^{\text{lim}}\). 
Similarly, the orientations of the hip joints \(\text{hip}_{l}\), and the knee joints \(\text{knee}_{l}\) (where \(l\) indicates the leg index), are bounded by limits \(\text{hip}^{\text{lim}}\) and \(\text{knee}^{\text{lim}}\), ensuring a consistent posture during movement.

\paragraph{Gait Constraints}  
These constraints define the robot's locomotion gait. 
The walking constraint ensures that only one foot is in contact with the ground at any given time, expressed as \(| n_{\text{foot contact}} - 1 |\), where \(n_{\text{foot contact}}\) denotes the number of feet in contact with the ground. 
For jumping, the constraint enforces that either zero or two feet are in contact with the ground, represented by \(| n_{\text{foot contact}} \bmod 2 |\). 
Foot air-time constraints ensure that each foot stays in the air for a sufficient amount of time during each locomotion cycle, defined by \( t_{\text{air time}}^{\text{des}} - t_{\text{air time}_j} \), where \(t_{\text{air time}_j}\) is the actual time that foot \(j\) is in the air.

Switching between walking and jumping behaviors requires retraining the network due to the mutually exclusive nature of the constraints.

\begin{table}
    \vspace{4pt}
	\centering
	\caption{Constraints applied during training. A constraint is considered violated when its associated cost \( c_i > 0 \).}
	\label{tab:constraints-bolt-rl}
	\resizebox{\columnwidth}{!}{
		\begin{tabular}{cc}
			\hline
			\multicolumn{2}{c}{\textbf{Safety Constraints}}                                                                                                  \\ \hline
			\multicolumn{1}{c|}{Knee or base collision}              & $c_{\text{knee/base contact}} = \mathbbm{1}_{\text{knee/base contact}}$                     \\
			\multicolumn{1}{c|}{Upside-down state}                   & $c_{\text{upsidedown}} = \mathbbm{1}_{\text{upsidedown}}$                                   \\
			\multicolumn{1}{c|}{Foot contact force limit}            & $c_{\text{foot contact}_j}=\left\|f^{\text{foot}_j}\right\|_2 - f^{\text{lim}}$       \\
			\multicolumn{1}{c|}{Torque limits}                       & $c_{\text{torque}_k}=\left|\tau_k\right| - \tau^{\text{lim}}$                           \\
			\multicolumn{1}{c|}{Joint velocity limits}               & $c_{\text{joint velocity}_k}=\left|\dot{q}_k\right| - \dot{q}^{\text{lim}}$             \\
			\multicolumn{1}{c|}{Joint acceleration limits}           & $c_{\text{joint acceleration}_k}=\left|\ddot{q}_k\right| - \ddot{q}^{\text{lim}}$       \\ \hline
			\multicolumn{2}{c}{\textbf{Posture Constraints}}                                                                                                   \\ \hline
			\multicolumn{1}{c|}{Base orientation}                    & $c_{\text{ori}}=\| \text{base ori}_{\mathrm{xy}} \|_2 - \text{base}^{\text{lim}}$ \\
			\multicolumn{1}{c|}{Hip orientation}                     & $c_{\text{hip}_l}=\mid \text{hip ori}_l \mid - \text{hip}^{\text{lim}}$           \\
			\multicolumn{1}{c|}{Knee orientation}                    & $c_{\text{knee}_l}=\mid \text{knee ori}_l \mid - \text{knee}^{\text{lim}}$        \\ \hline
			\multicolumn{2}{c}{\textbf{Gait Constraints}}                                                                                                     \\ \hline
			\multicolumn{1}{c|}{Walking (single foot contact)}       & $c_{\text{foot contact}} = | n_{\text{foot contact}} - 1 |$                                 \\
			\multicolumn{1}{c|}{Jumping (alternating zero or two foot contacts)} & $c_{\text{foot contact}} = | n_{\text{foot contact}} \bmod 2 |$                              \\
			\multicolumn{1}{c|}{Foot air time}                       & $c_{\text{air time}_j}=t_{\text{air time}}^{\text{des}} - t_{\text{air time}_j}$            \\ \hline
		\end{tabular}
	}
\end{table}

\subsection{Domain Randomization}

To ensure robust zero-shot sim-to-real transfer, we employ domain randomization during training, as summarized in Table~\ref{tab:domain_randomization}. 
We introduce variations in both the environment and the robot's dynamics, training the policy to generalize better to real-world uncertainties. 
Additionally, noise is injected into the observations to simulate sensor imperfections and further improve robustness.

\paragraph{Environment and Dynamics Randomization}  
First, we randomize terrain properties by varying the ground friction coefficient and introducing uneven terrain profiles. 
The robot's dynamics is also randomized by modifying joint friction, shifting the center of mass (CoM) of the base, and perturbing the masses and inertia tensors of the robot's links. 
By randomizing the physics parameters, we prevent the policy from overfitting to both the simulator and the specific URDF model of the robot, ensuring it remains robust to real-world variations rather than exploiting simulator-specific artifacts.

\paragraph{State and Command Perturbations}  
To prevent overfitting to a fixed initial condition, the robot's initial state is randomized by applying random offsets to joint positions and velocities at spawn. External forces are randomly applied to the base to simulate real-world disturbances, and additional latency effects are introduced by simulating actuator delays and command variations.  

\paragraph{Observation Noise}  
To account for sensor imperfections, noise is added to the observations. 
This noise injection further enhances robustness by preventing the policy from relying on overly precise state estimates.  

\subsection{Key Factors for Sim-to-Real Transfer}  
Among all randomization parameters, joint friction variations and actuator delays are particularly critical for successful sim-to-real transfer. 
Without accurately modeling joint friction, the robot fails to maintain balance and collapses immediately upon deployment. 
Likewise, actuator delays introduce inherent latency that must be accounted for to ensure smooth and stable motion execution. 
While other randomization parameters enhance overall robustness, these two factors are found to be indispensable for achieving reliable real-world locomotion.

\begin{table}
    \centering
    \vspace{4pt}
    \caption{Domain randomization parameters for training. Key sim-to-real factors are in bold.}
    \label{tab:domain_randomization}
    \renewcommand{\arraystretch}{1.3}
    \resizebox{\columnwidth}{!}{
    \begin{tabular}{lll}
        \hline
        \textbf{Category} & \textbf{Parameter} & \textbf{Distribution} \\ \hline

        \multirow{2}{*}{\textbf{Terrain}} 
            & Ground friction coefficient & \( \mathcal{U}(0.4,\ 1.5) \) \\
            & Height noise & \( \mathcal{U}(0.0,\ 0.02) \) m, step = 0.005 m \\ \hline

        \multirow{4}{*}{\textbf{Robot Dynamics}} 
            & Mass scale factor & \( \mathcal{U}(0.8,\ 1.2) \) \\
            & Inertia scale factor & \( \mathcal{U}(0.8,\ 1.2) \) \\
            & Base CoM displacement & \( \mathcal{U}(-0.02,\ 0.02) \) m \\
            & \textbf{Joint friction coefficient} & \( \mathcal{U}(0.01,\ 0.1) \) \\ \hline

        \multirow{3}{*}{\textbf{Initial State}} 
            & Base position (x, y) & \( \mathcal{U}(-0.5,\ 0.5) \) m \\
            & Base yaw angle & \( \mathcal{U}(-\pi,\ \pi) \) rad \\
            & Base linear velocity & \( \mathcal{U}(-0.3,\ 0.3) \) m/s \\
            & Base angular velocity & \( \mathcal{U}(-0.1,\ 0.1) \) rad/s \\
            & Joint pos./vel. scale factor & \( \mathcal{U}(0.9,\ 1.1) \) \\ \hline

        \multirow{4}{*}{\textbf{Events}} 
            & \textbf{Actuation delay} & \( \mathcal{U}(0,\ 2) \) simulation steps \\
            & Push linear velocity (x, y) & \( \mathcal{U}(-0.5,\ 0.5) \) m/s \\ 
			& Push linear velocity (z) & \( \mathcal{U}(-0.1,\ 0.1) \) m/s \\ 
			& Push angular velocity & \( \mathcal{U}(-0.5,\ 0.5) \) rad/s \\ \hline

        \multirow{4}{*}{\textbf{Observation Noise}} 
            & Base angular velocity & \( \mathcal{U}(-0.2,\ 0.2) \) rad/s \\
            & Projected gravity noise & \( \mathcal{N}(0,\ 0.05^2) \) with bias \( \mathcal{U}(0,\ 0.05) \) \\
            & Joint position noise & \( \mathcal{U}(-0.01,\ 0.01) \) rad with bias \( \mathcal{U}(0,\ 0.05) \) rad \\
            & Joint velocity noise & \( \mathcal{U}(-1.5,\ 1.5) \) rad/s \\ \hline
    \end{tabular}
    }
\end{table}

\section{Experiments}
\label{sec:experiments}
We conducted quantitative and qualitative experiments on the real robot to evaluate our controller, highlighting its strengths and identifying potential limitations.

\subsection{Experimental Setup}
The policies were trained using the IsaacLab framework \cite{mittal2023orbit}, leveraging the PPO implementation from the CleanRL library \cite{huang2022cleanrl, schulman2017proximal}. 
Our approach is implemented within our \textit{CaT} framework \cite{chane2024cat}, available on GitHub\footnote{\url{https://gepetto.github.io/BoltLocomotion/}}\label{fn:github}. 

After successful training in simulation, the learned policies were directly deployed on the real Bolt bipedal robot, described in Sec. \ref{sec:hardware}.
The policy runs at a frequency of 50Hz on an Apple Mac M3 Max CPU. 
Target joint positions are transmitted to the onboard PD controller, operating at a frequency of 10kHz.
The complete inference pipeline, including code and logs, is available on GitHub\footnotemark[\value{footnote}] for reproducibility.

\subsection{Evaluation Metrics}
To quantitatively evaluate the performance of our approach, we introduce distinct evaluation metrics across three axes:
\begin{itemize}
	\item \textbf{Performance Axis:} These metrics evaluate the robot's maximum velocity and the accuracy of its velocity control.
	\item \textbf{Autonomy Axis:} These metrics evaluate the short-term autonomy by monitoring power consumption, and long-term autonomy based on the ground reaction force computed for each foot.
	\item \textbf{Robustness Axis:} These metrics evaluate the controller's capacity to reject external disturbances.
\end{itemize}
Additionally, we conducted slip recovery and jumping experiments to further demonstrate the robustness and adaptability of the controller.

\subsubsection{Performance Metrics}
\paragraph{Velocity Accuracy}
To assess the accuracy of the velocity tracking of our controller, we computed the steady-state error for a set of velocity references along either the $x$-axis and the $y$-axis.
The velocity tracking error $\bar{\boldsymbol{\epsilon}}$ is defined as the difference between the reference velocity $\mathbf{v}^*$ and the robot's mean steady-state velocity $\bar{\mathbf{v}}$.

\paragraph{Maximum Velocity}
The maximum velocity is defined as the highest steady-state mean velocity attained along the $x$ or the $y$-axis, in both directions, as the velocity reference is progressively increased.
The process continues until the robot ceases to accelerate.

\subsubsection{Autonomy Metrics}
\paragraph{Short-Term Autonomy}
To assess the short-term autonomy of our controller, we measure the CoT for different velocity references along the $x$ and $y$-axes. 
The CoT is defined as:
\begin{equation}
	\text{CoT} = \frac{\sum_{i=0}^{N-1} \sum_{k=0}^{n_{\boldsymbol{u}}-1} P^{\text{loss}}_{i,k}}{N m g \left\| \bar{\mathbf{v}} \right\|_2}
\end{equation}
where $N$ is the total number of time steps during the experiment, $n_{\boldsymbol{u}}$ is the number of joints, $m$ is the robot's mass, $g$ is the gravitational acceleration, and $\bar{\mathbf{v}}$ denotes the robot's average velocity throughout the experiment. 
The power loss $P^{\text{loss}}_{i,k}$ for actuator $k$ at timestep $i$ is:
\begin{equation}
	P^{\text{loss}}_{i,k} = P^J_{i,k} + P^f_{i,k}
\end{equation}
where $P^J_{i,k}$ and $P^f_{i,k}$ are the Joule and friction power losses, respectively, as described by Fadini et al. \cite{9560988}.

\paragraph{Long-Term Autonomy}
To evaluate the potential damage caused by foot impacts on the ground, we estimate the ground reaction force using the following equation of motion:
\begin{equation}
	M(\mathbf{q}) \ddot{\mathbf{q}} + b(\mathbf{q}, \dot{\mathbf{q}}) = \tau(\mathbf{u}) + J_c(\mathbf{q})^T f
\end{equation}
where $\mathbf{q}$ and $\dot{\mathbf{q}}$ represent the measured joint positions and velocities, respectively, $\tau$ denotes the joint torques generated by the control inputs $\mathbf{u}$, $J_c$ is the contact Jacobian of the feet, $\ddot{\mathbf{q}}$ is the joint acceleration, and $f$ is the ground reaction force.

This equation is solved using well-established algorithms from the literature \cite{carlosmastalliCrocoddylEfficientVersatile2019, justincarpentierProximalSparseResolution2021}.
The computed ground reaction forces are then analyzed as part of our benchmark to assess impact-related wear and long-term autonomy.

\subsubsection{Robustness Criteria}
\paragraph{Push Recovery}
To quantify the robot's ability to reject external disturbances, an instrumented stick is used to apply controlled pushes to the robot's base.
The pose of the stick and the force exerted on the robot are measured and the applied force is progressively increased in different directions until the robot loses stability and falls.
The maximum impulse sustained before falling is recorded.
The impulse is computed as the discrete integral:
\begin{equation}
	\lambda = \delta t \sum_{i=1}^{M-1} \frac{f_{i} + f_{i-1}}{2}
\end{equation}
where $M$ denotes the total number of time steps during the push event, $f_i$ represents the measured force at time step $i$, and $\delta t$ is the time step duration.

\subsection{Results}
\subsubsection{Performance Criteria}
\begin{figure}
	\centering
	\begin{subfigure}{0.49\linewidth}
		\includegraphics[width=\textwidth]{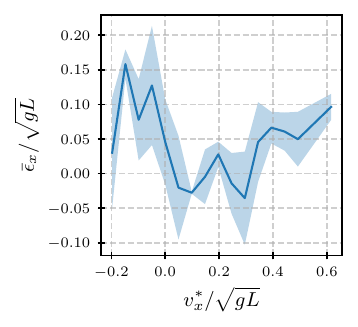}
	\end{subfigure}
	\begin{subfigure}{0.49\linewidth}
		\includegraphics[width=\textwidth]{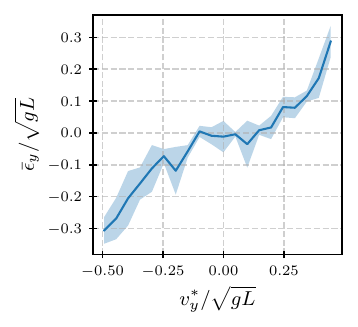}
	\end{subfigure}
	\caption{Normalized steady-state velocity error versus normalized velocity command along the $x$-axis (left) and the $y$-axis (right). The solid line represents the mean, and the shaded area indicates the standard deviation.}
	\label{fig:vel_acc}
\end{figure}

\paragraph{Velocity Accuracy}
Velocity commands range from $-1$ m/s to $1$ m/s along the $x$ and $y$-axes separately, with increments of $0.1$ m/s.
Each experiment is repeated at least three times per velocity reference.
The steady-state velocity is determined by applying a moving average filter and extracting the stable segment of the signal.
Results are presented in Fig.~\ref{fig:vel_acc}, where all velocities are normalized by $\sqrt{gL}$ (with $L$ being the robot's leg length), corresponding to the Froude number normalization, as outlined in \cite{locomotion2003}. 
This normalization enables meaningful comparisons across different locomotion speeds and morphologies.

Overall, the robot successfully moves in the commanded direction (e.g., forward when instructed to do so).
Along the $x$-axis, it achieves significantly higher forward velocities than backward ones but exhibits considerable drift, leading to poor velocity tracking. 
In contrast, along the $y$-axis, it demonstrates symmetric performance in both left and right directions.
These results confirm that the robot is not only capable of balancing but also of following velocity commands. 
However, drift becomes more pronounced at higher velocity references. 
This drift can be attributed to the absence of linear velocity measurements in our setup, which eliminates direct feedback necessary for accurate reference tracking. 
As a result, the robot lacks immediate velocity feedback, making it harder to adjust movements in real-time for accurate control.
This issue is compounded by the robot's inherently unstable morphology, further amplifying the challenge of maintaining precise control at higher speeds.

\paragraph{Maximum Velocity}
Velocity references along the desired axis increase in increments of $0.1$ m/s until the robot falls.
The maximum steady-state velocity achieved is recorded.
Table \ref{tab:velocities} summarizes the resulting performance, with all velocities normalized as before.

\begin{table}
	\centering
	\caption{Minimum and maximum normalized velocities achieved along the $x$ and $y$ axes.}
	\label{tab:velocities}
	\begin{tabular}{lcc}
		\toprule
		         & $v_{\text{min}}/\sqrt{gL}$ & $v_{\text{max}}/\sqrt{gL}$ \\
		\midrule
		$x$-axis & -0.34                      & 0.54                       \\
		$y$-axis & -0.31                      & 0.30                       \\
		\bottomrule
	\end{tabular}
\end{table}
It is worth emphasizing that the robot achieves higher velocities in forward motion compared to backward, while exhibiting symmetric performance in lateral directions.
When walking forward, the robot achieves a Froude number normalization of \( Fr = 0.54 \), which is comparable to the typical human walking value of \( Fr \simeq 0.5 \). 
Despite lower maximum velocities in the backward and lateral directions, the robot's performance remains competitive. 
Qualitative observations from video footage indicate that prior work \cite{9361296} resulted in significantly lower velocities, highlighting the improved agility of our approach.
\subsubsection{Autonomy Criteria}
\begin{figure}
	\centering
	\includegraphics[width=\linewidth]{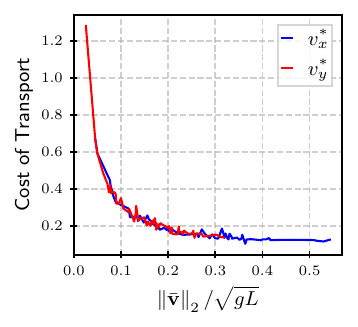}
	\caption{Cost of transport as a function of the steady-state velocity norm. Blue represents velocity references along the $x$-axis, while red corresponds to those along the $y$-axis.}
	\label{fig:cost_of_transport}
\end{figure}

\paragraph{Short-Term Autonomy}
The CoT is evaluated at various steady-state velocities of the robot.
Results are categorized based on the velocity command direction (forward/backward or left/right) and are presented in Fig.~\ref{fig:cost_of_transport}.

For velocity references along both the $x$ and $y$-axes, the CoT decreases as the steady-state velocity norm increases. 
It can be used in future work to model battery consumption for an embedded version of the robot. 
Additionally, it establishes a useful benchmark for comparison in future studies.

\paragraph{Long-Term Autonomy}
The mean ground reaction force (GRF) is found to be independent of the robot's velocity, remaining at $90\% \pm 6.2\%$ of the robot's weight. 
This value being below 100\% is explained by the presence of short double support phases—periods during which both feet briefly share the load. 
Since the robot primarily walks with single-foot support, the average GRF per foot remains below full body weight. 
This indicates reduced GRF during gait transitions, which contributes to mechanical longevity.
Furthermore, it confirms that the contact force constraint imposed during training is maintained in practice.

\subsubsection{Robustness Criteria}
\begin{figure}
	\centering
	\includegraphics[width=\linewidth]{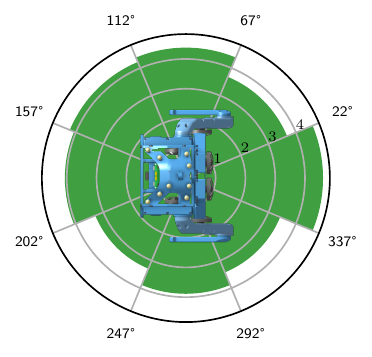}
	\caption{Maximum impulse before falling (expressed as \% of weight $\times$ seconds) as a function of the push angle, where $0^\circ$ corresponds to a push applied to the back of the robot's base.}
	\label{fig:push_recovery}
\end{figure}

\paragraph{Push Recovery}
The robot was subjected to over 300 pushes, covering a full range of directions around its base.
Fig.~\ref{fig:push_recovery} presents the maximum impulse, normalized as a percentage of the robot's weight, that the robot could withstand without falling, categorized in $45^\circ$ quadrants.

Overall, the robot exhibits greater robustness to recover from pushes applied to the back and sides.
Conversely, it is generally less robust when pushed from the front or at the corners, except for the upper right corner.
This anomaly can be attributed to insufficient data points recorded in that region.
For comparison, an impulse of $4\%$ of the robot's weight over one second is equivalent to a $70$ kg human being struck by a $2.8$ kg ball over the course of $1$ second, demonstrating the robustness of our method.

\subsection{Additional Results}
\begin{figure*}
    \vspace{4pt}
	\centering
	\includegraphics[width=0.8\linewidth]{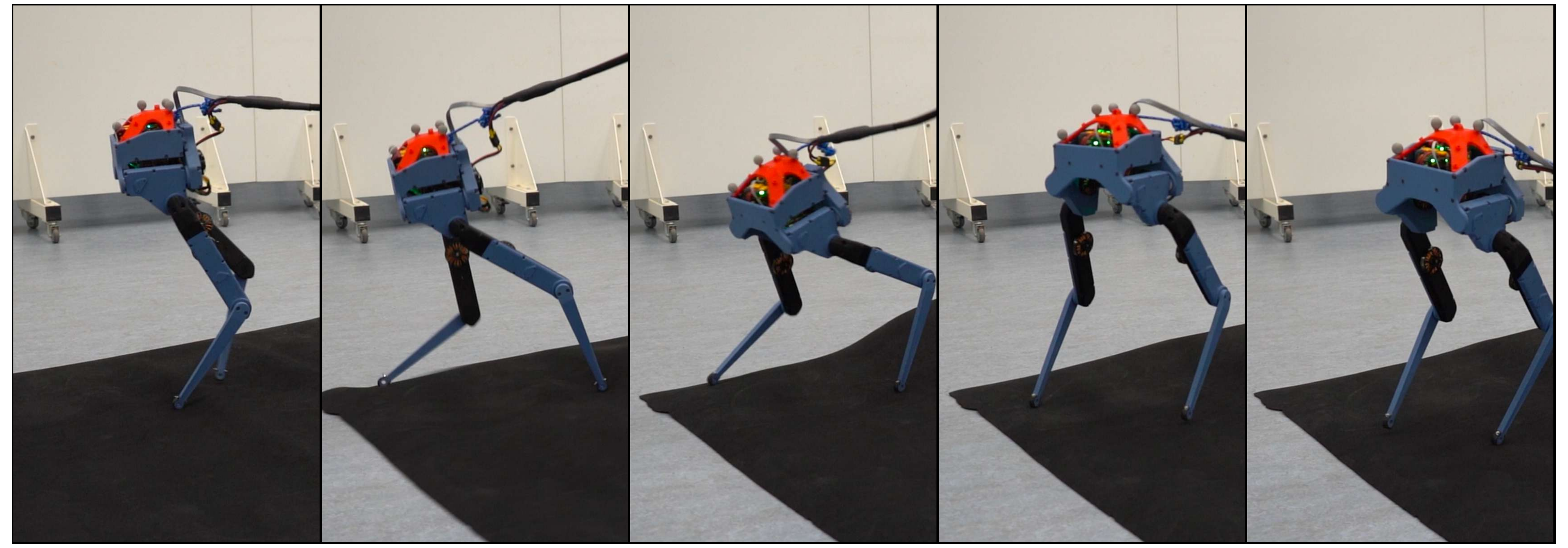}
	\caption{Motion capture sequence of the robot recovering from a slip after the carpet was pulled away. Captured at 20 Hz.}
	\label{fig:slip_recovery}
\end{figure*}

To qualitatively demonstrate the robustness of the balancing process, we present additional results. 
Fig.~\ref{fig:slip_recovery} shows the robot recovering from a slip after the carpet was pulled out while it was walking on it. 
These images are extracted directly from the accompanying video, which provides a more dynamic visualization of the entire recovery sequence.

\begin{figure*}
	\centering
	\includegraphics[width=0.8\linewidth]{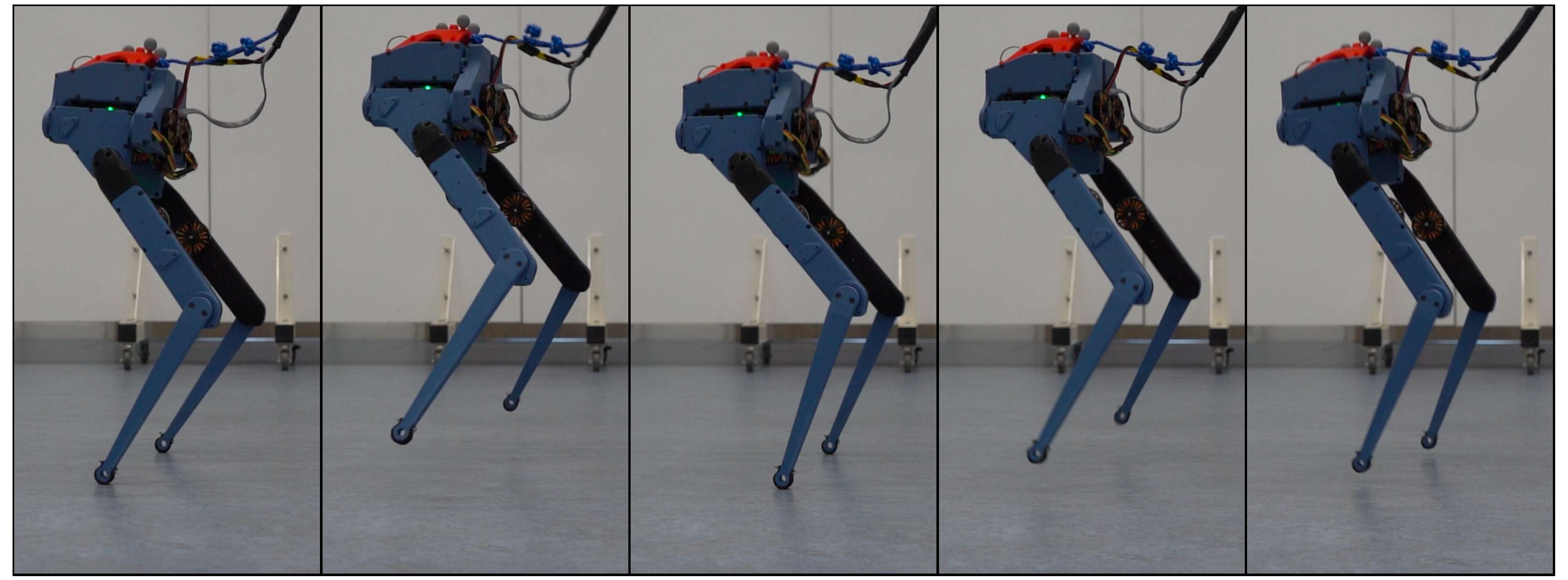}
	\caption{Motion capture of the robot jumping. Captured at 33 Hz.}
	\label{fig:jumps}
\end{figure*}

Fig.~\ref{fig:jumps} shows the robot performing jumps, enabled by modifying the gait constraint to allow ballistic phases (see Sec.\ref{subsec:constraints}).
It successfully completed over 50 consecutive jumps, maintaining this behavior for more than 15 seconds, as demonstrated in the accompanying video.

\section{Conclusion}
\label{sec:conclusion}
This work explored the control of a point-foot bipedal robot, emphasizing its significance in achieving anthropomorphic locomotion. 
Unlike flat-footed bipeds, point-foot robots face unique stability challenges, making their control an interesting research topic.  

We developed a constrained RL approach and successfully transferred policies from simulation to the real robot without additional fine-tuning. 
Our experiments demonstrated that the robot can balance, track velocity commands, jump, and resist external disturbances, showcasing a high level of robustness.  

Despite these achievements, real-world deployment remains challenging due to hardware fragility and external factors, such as the use of a wire connecting the robot to an external computer. 
This setup, while necessary for communication and data processing, limits mobility and introduces additional complexities during operation.
Future work will focus on refining robustness, exploring additional locomotion behaviors, and further improving real-world navigation capabilities.  

\section*{Aknowledgment}
We thank the Fablaas and Electronics teams for their support in robot repairs, and the biomechanical team for constructing the force measurement stick. 
The Bolt platform was supported by ROBOTEX 2.0 (Grants ANR-10-EQPX-44-01, TIRREX-ANR-21-ESRE-0015) and AS2 (ANR-22-EXOD-0006), and ANR-19-PI3A-0004. 
This work was granted access to the HPC resources of IDRIS under the allocation 2025-AD011016104 made by GENCI.

\bibliographystyle{IEEEtran}
\bibliography{bib}

\end{document}